# Bipedal Locomotion Optimization by Exploitation of the Full Dynamics in DCM Trajectory Planning

Amirhosein Vedadi
Student, Center of Advanced Systems
and Technologies (CAST),
School of Mechanical Engineering,
College of Engineering,
University of Tehran, Tehran, Iran.
amirhosein.vedadi@ut.ac.ir

Kasra Sinaei
Student, Center of Advanced Systems
and Technologies (CAST),
School of Mechanical Engineering,
College of Engineering,
University of Tehran, Tehran, Iran.
kasra.sinaei@ut.ac.ir

Pezhman Abdolahnezhad
Student, Center of Advanced Systems
and Technologies (CAST),
School of Mechanical Engineering,
College of Engineering,
University of Tehran, Tehran, Iran.
abdolahnezhad.pe@ut.ac.ir

Shahriar Sheikh Aboumasoudi
Student, Center of Advanced Systems
and Technologies (CAST),
School of Mechanical Engineering,
College of Engineering,
University of Tehran, Tehran, Iran.
shahriar.masoudi@ut.ac.ir

Aghil Yousefi-Koma
Professor, Center of Advanced Systems
and Technologies (CAST),
School of Mechanical Engineering,
College of Engineering,
University of Tehran, Tehran, Iran.
aykoma@ut.ac.ir

*Abstract*—Walking motion planning based on Divergent Component of Motion (DCM) and Linear Inverted Pendulum Model (LIPM) is one of the alternatives that could be implemented to generate online humanoid robot gait trajectories. This algorithm requires different parameters to be adjusted. Herein, we developed a framework to attain optimal parameters to achieve a stable and energy-efficient trajectory for real robot's gait. To find the optimal trajectory, four cost functions representing energy consumption, the sum of joints velocity and applied torque at each lower limb joint of the robot, and a cost function based on the Zero Moment Point (ZMP) stability criterion were considered. Genetic algorithm was employed in the framework to optimize each of these cost functions. Although the trajectory planning was done with the help of the simplified model, the values of each cost function were obtained by considering the full dynamics model and foot-ground contact model in Bullet physics engine simulator. The results of this optimization yield that walking with the most stability and walking in the most efficient way are in contrast with each other. Therefore, in another attempt, multi-objective optimization for ZMP and energy cost functions at three different speeds was performed. Finally, we compared the designed trajectory, which was generated using optimal parameters, with the simulation results in Choreonoid simulator.

*Keywords—humanoid robot, trajectory planning, contact model, linear inverted pendulum model, divergent component of motion, multi-objective optimization, genetic algorithm*

## I. Introduction

Humanoid robots have always been of interest to robotic researchers due to their ability to operate in human-specific environments. However, the humanoid robot's high degrees of freedom and unstable nature cause many challenges for designing a stable and optimal trajectory. Vukobratovic et al. introduced the Zero Moment Point (ZMP) criterion [1], [2], which is used as a measure to evaluate the stability of the humanoid robot. ZMP is the point at which the horizontal torques due to ground reaction forces are zero, and as long as this point is in the support polygon of the robot, no torque is created about the edges of the robot's sole and the robot's movement will be stable. After introducing this criterion, various methods were proposed for motion planning and control using this concept.

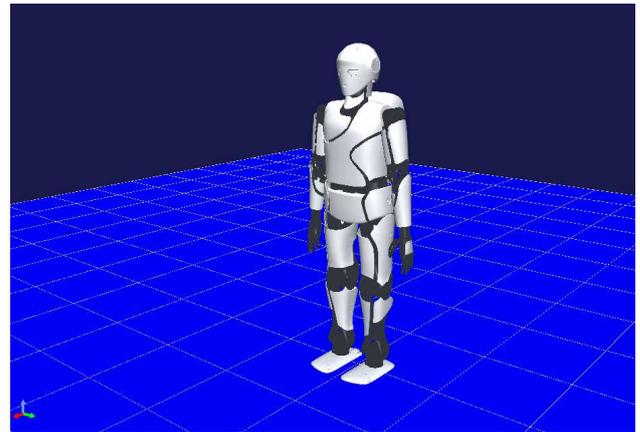

Fig. 1. Surena IV humanoid robot in Choreonoid simulator

There are two major approaches to motion planning of bipedal robots. The first approach relies on the simplified dynamic model of robot while the other approach considers the full dynamics model of robot. In the first approach, Kajita introduced Linear Inverted Pendulum Model (LIPM) [3]. Due to the simplicity and linearity of this model, it is a versatile tool for online motion planning and control. Nevertheless, this simplicity will come with a modeling error. Other models have been proposed to reduce this error. One of these models is an inverted pendulum with three masses [4], [5]. Guo et al. discussed a new gait planning and generation using a 3D force-pattern-based inverted pendulum model where the reaction force of the support leg is defined by a parametric function whose parameters are optimized using a nonlinear optimizer. Then by employing Runge-Kutta solver, a stable Center of Mass trajectory at a given constant speed is generated [6]. To consider the torque around the robot center of mass, the inverted pendulum model with the flywheel has been used [7].

Pratt et al. introduced the concept of Capture Point (CP) and used it for push recovery [7]. CP is the Divergent Component of Motion (DCM) in LIPM. CP is the point at which the robot comes to a stop if it steps on it. Takenada et al. Used the DCM concept for trajectory planning of the Asimo robot [5]. Englsberger et al. also used DCM for trajectory planning and control of the humanoid and extended its concept to three dimensions [8], [9]. Shafiee et al. presented a multi-layer DCM-based push recovery Controller [10], [11],[12]. Tian et al. introduced Enhanced Capture Point (ECP) and used it for trajectory planning, and developed a controller based on it [13]. Kim et al. proposed a method for stabilizing dynamic walking of a humanoid by optimizing capture point trajectory. The algorithm used for achieving this purpose was Particle Swarm Optimization (PSO) [14].

As mentioned before there exists another method for motion planning which considers full body dynamics and is employed by many researchers. Kumar Yadav et al. proposed a general method of walking trajectory generation. Their approach employs sinusoidal function, which was used to predict hip joint trajectory and cubic spline to generate knee and ankle joint trajectories. Genetic Algorithm (GA) was used to obtain optimal waypoints for these functions [15]. Huan et al. proposed an innovative method to optimize humanoids' trajectory generation process that allows them to walk stably. In their research, four humanoid trajectory generation design parameters were considered and optimized using Jaya optimization algorithm. They defined two objective functions, one of which represented humanoid stability and the other represented distance between foot-lift reference height and foot-lift actual height. The results were tested on HUBoT-4 biped robot, and a stable walk was obtained [16]. Sadedel et al. [17] and khadiv et al. [18] also adopted full dynamics approach to trajectory generation. A gait planning process was presented in their work, and its parameters were optimized using genetic algorithms for different objective functions. Finally, results were approved through testing them on the Surena III humanoid. In [19], trajectory generation of the Surena IV robot was based on the parameterization of the task space by polynomials to guarantee the smooth motion of the robot. These parameters were optimized using the ZMP cost function derivom the full dynamics model.

In this paper, we generate the Center of Mass (CoM) trajectory of the Surena IV robot using the LIPM and the DCM concept. With the help of this algorithm, the robot's trajectory could be designed online, and higher speeds could be attained in walking. In this algorithm, several parameters need to be set for trajectory planning. By changing these parameters, different trajectories with different stability characteristics and energy levels are obtained. In order to be able to use appropriate parameters for online trajectory planning, we developed a framework for optimization with various objective functions. Although the trajectory planning in this paper is based on the LIPM, in order to reduce modeling errors, we obtained the measured parameters for optimization with the help of Bullet physics engine, in which the full dynamics model of the robot has been considered. This means that the foot-ground contact has been modeled as well.

This paper continues as follows: Section II examines the DCM-based trajectory generation for the robot. The third section details the optimization problem. Section IV discusses the results of the simulation on Surena IV. Finally, in the fifth section, the discussions are concluded.

## II. DCM Based Trajectory Generation

### A. LIP Model and DCM Concept

To estimate the dynamics of the humanoid robot using the LIPM, it is assumed that the total mass of the robot is concentrated in a point equivalent to the robot's CoM, and the height of the tobot is constant ($z_0$) which is an assumption that makes the differential equation of the robot dynamics linear. This concentrated mass is connected to the torque-free base joint of the pendulum which is equivalent to the ZMP in the humanoid robots. With these assumptions, the system equation is as follows:

$$\ddot{x} = \omega^2(x - r_{zmp}) \qquad (1)$$

where $x$ is the position of CoM of the robot, $r_{zmp}$ is the position of ZMP and $\omega = \sqrt{g/z_0}$ is the model natural frequency. The equation in the y direction is identical to (1). By considering a state variable such as $\xi$, the second-order unstable dynamics of (1) becomes the following two first-order equations:

$$\dot{\xi} = \omega(\xi - r_{zmp}) \qquad (2)$$

$$\dot{x} = -\omega(x - \xi) \qquad (3)$$

(2) is unstable and that is why $\xi$ is called Divergent Component of Motion (DCM). Contrary to (2), (3) is stable and the $x$ always converges to $\xi$. Therefore, having a suitable trajectory for $\xi$, the trajectory of the CoM is obtained from (2).

### B. Trajectory Generation

In this section, we used the method presented in Reference [9] to generate robot trajectories. By integrating (2), the DCM's time domain equation is obtained as follows:

$$\xi(t) = r_{zmp} + e^{\sqrt{g/z_0}t}(\xi_{init} - r_{zmp}) \qquad (4)$$

where $\xi_{init}$ is initial DCM, $t$ is the time that each step lasts and the $z_0$ is the robot's height, which is a constant value. With the help of (4), it is possible to generate the trajectory of DCM in each step. According to the robot step planning, the ZMP location is assumed to be in the center of each footprint. Also, to get the initial DCM in each step, it is assumed that DCM matches the ZMP in the last step, and the previous DCM positions are calculated recursively according to (5):

$$\xi_{init}^i = \xi_{end}^{i-1} = r_{zmp}^i + e^{-\sqrt{g/z_0}t_{step}}(\xi_{end}^i - r_{zmp}^i) \quad (5)$$

$\xi_{init}^i$ and $\xi_{end}^i$ are the i[th] step's initial and final DCM respectively, and $t_{step}$ is the time for each step. Now, with the help of these values and by using (4), the DCM trajectory could be generated in the single support phase, as illustrated in Fig. 2.

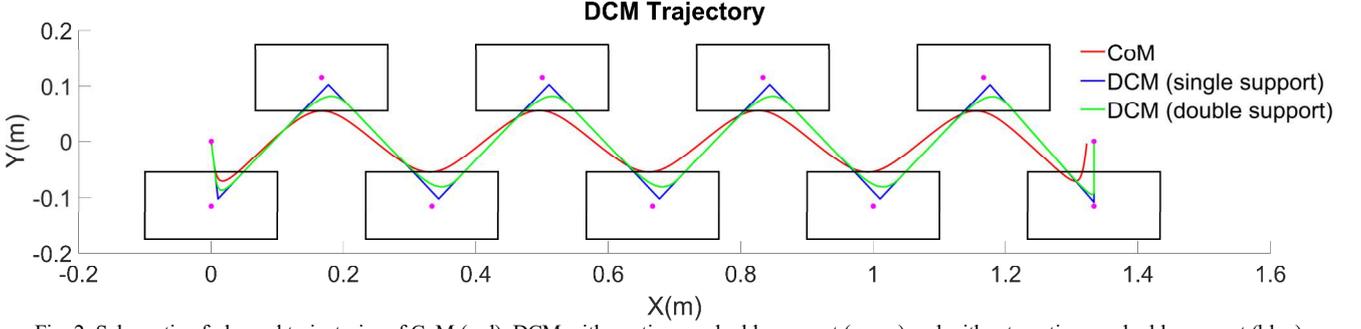

Fig. 2. Schematic of planned trajectories of CoM (red), DCM with continuous double support (green) and without continuous double support (blue)

Until now, it has been assumed that the robot's support foot transition occurs instantly, and there is no double support phase while walking. With this assumption, the external forces acting on the robot become discontinuous in the moments of supporting foot changes [9]. To improve the desired DCM trajectory and eliminate the sharp corners of the trajectory, we used a 3rd degree polynomial to generate a double support phase trajectory. Initial and final points of double support phase trajectory are obtained using (4).

$$\xi_{init,DS}^i = r_{zmp}^{i-1} + e^{-\sqrt{g/z_0}(\Delta t_{init,DS})}(\xi_{init}^i - r_{zmp}^{i-1}) \quad (6)$$

$$\xi_{end,DS}^i = r_{zmp}^i + e^{\sqrt{g/z_0}(\Delta t_{end,DS})}(\xi_{init}^i - r_{zmp}^i) \quad (7)$$

In (6) and (7), $\Delta t_{init,DS} = \alpha \cdot t_{DS}$ and $\Delta t_{end,DS} = (1 - \alpha) \cdot t_{DS}$, where $\Delta t_{init,DS}$ and $\Delta t_{end,DS}$ are the time intervals before and after instantaneous supporting foot transition, respectively. The Final generated DCM trajectory is shown in Fig. 2. As mentioned before, CoM trajectory could be obtained by integrating (3) over time:

$$x(t) = e^{-\sqrt{g/z_0}t}(x(0) + \sqrt{\frac{g}{z_0}} \int e^{\sqrt{g/z_0}t} \xi(t) dt) \quad (8)$$

In order for the robot to be able to walk, ankle trajectories must be generated as well. To this end, we utilized a 5th degree polynomial and maximum height, initial and final positions, and their corresponding velocities were considered as boundary conditions.

III. OPTIMIZATION PROCEDURE

A. Optimization Problem

Trajectory planning with the help of DCM has various parameters that a change in each of them alters the stability and dynamic conditions of the robot. The main goal of this paper is to find optimal parameters so that by using them in designing the robot's online trajectory, a stable and optimal trajectory can be designed. Also, this trajectory should be such that it acknowledges the robot's mechanical limitations. Some of these limitations are due to dynamic constraints such as motors' maximum torque and motors' maximum velocity. In addition to that, there are also robot's kinematic constraints [20]. Although the Geometric Inverse Kinematic method introduced by Kajita [21] can handle singularities, the full body motion generated under some conditions may lead to unstable walking behaviors. As a result, some constraints should be applied to the optimization problem, and the search area should be limited.

The developed framework in this work optimizes parameters of walking pattern generator: α which is used for double support timing, $r_{DS} = {t_{DS}}/{t_{SS}}$ which is the ratio of the double support phase time to the total step time, $t_{step}$ which is step duration, $z_0$ which is pelvis height, and $h_{ankle}$ which is the maximum ankle swing height. robot walking average speed is set at the beginning of the simulation and will stay constant. Since the robot's speed is kept at a constant value during optimization, step length could be calculated from $t_{step}$.

Objective functions values will be calculated during a limited time (5 seconds) of continuous walking on a straight line by considering the full dynamics and foot-ground contact models in PyBullet Simulation.

B. Objective Functions and Constraints

Humanoid robots are known for their poor energy efficiency since they use servo motors as actuators, which require high current. Overall, energy consumption is one of the most critical criteria based on which we derive our first objective function. As expressed in (9) it is simply sum of the energy consumed in all the lower limb joints.

$$J_E = -\sum_{i=1}^{12} E_i \quad (9)$$

LIPM guarantees the stability of the robot during dynamic walking; however, ZMP of the robot while following the offline pattern generated from LIPM model may vary slightly from what we expected. So we do consider its stability regarding its distance from the support polygon's edges. Based on the ZMP stability criterion, we define the second objective function.

$$\begin{cases} J_{ZMP} = -r(p_{zmp}, V) & \text{if ZMP is inside} \\ J_{ZMP} = r(p_{zmp}, V) & \text{if ZMP is not inside} \end{cases} \quad (10)$$

In (10), V is a set of points ($V_i \in R^3$) presenting corners of robot's soles, and $r(x, V)$ is the minimum distance of point x from a polygon, shaped from vertex of point in set V.

Since we are doing a minimization problem and we want the ZMP to be inside of the support polygon, we multiply its distance from the edge of the support polygon by -1 whenever it is inside. In order to decide whether the point is inside or outside of the support polygon we used the concept of windings number [22].

The following two cost functions are the sum of joints velocity and applied torque at each lower limb joint of the robot.

$$J_{torque} = -\sum_{i=1}^{12} T_i \quad (11)$$

$$J_{vel} = -\sum_{i=1}^{12} \dot{q}_i \quad (12)$$

In (11) and (12), $T_i$ is torque of i$^{th}$ joint, and $\dot{q}_i$ is velocity of i$^{th}$ joint in one step of simulation. Simulation frequency is kept constant at 240 Hz during simulations.

We consider three constraints for our optimization problem. The first one comes from the robot's joints workspaces that define each joint's range of motion. We impose our second constraint on actuators torque obtained from the robot's actuators maximum output torque. And last constraint confines the robot's center of mass height in order to eliminate those samples in the population that result in falling.

*C. Multi-objective and Single-objective Optimization*

With the ordinary GA method mentioned in the previous section, we can only optimize one of these objectives at a time; however, we seek a solution that optimizes the walking pattern trajectory considering all the mentioned criteria. Therefore, we are dealing with a Pareto optimization problem. In order to observe the effects of walking parameters on each item, we can use ordinary GA, but in order to obtain the best possible parameters, we use a NSGA-II algorithm introduced in 2002 by Deb [23]. This algorithm uses non-dominated sorting to rank members of population. After applying regular Genetic Operations over a number of iterations, it reaches Pareto front of the optimization problem. Then we can choose among the solutions in the Pareto front based on their distance from the utopia point (a fictitious solution that minimizes all objective functions, also called knee point). In this paper, we select the closest point of the Pareto front to the utopia point since both of our objective functions are of the same importance to us.

Increasing the dimension of objective functions space (number of objectives) will significantly increase computation time and require larger populations to reach a Pareto front set with acceptable accuracy.

According to our single objective optimization results which will be explained further in section IV.A, $J_E$ aligns with $J_{vel}$ and $J_{torque}$. So, we only consider $J_E$ and $J_{ZMP}$ in multi-objective optimization.

$$\min(J_{ZMP}(\vec{x}), J_E(\vec{x})) \quad (13)$$

$$\vec{x} = [\alpha, r_{DS}, t_{step}, z_0, h_{ankle}]$$

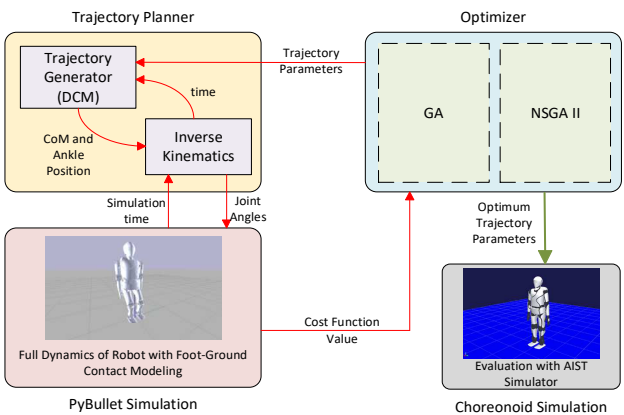

Fig. 3. Schematic of the developed framework

In (13) we can see the definition of our multi-objective problem. Having all these parameters, the optimizer node of our software can request trajectories for the robot's CoM and ankles, and the simulator node returns objective functions after executing the walking pattern based on received trajectories, considering the full dynamic model of the robot and the foot-ground contact model. NSGA-II and normal Genetic algorithm parameters were set based on the values presented in Table I.

In NSGA II selection operator creates a mating pool in which the next members of the next generation are chosen from the best solutions among both parents and offspring. Hence, the algorithm itself is inherently elitist.

TABLE I.  OPTIMIZATION ALGORITHMS PARAMETERS

| Algorithm | Population size | Crossover Type (prob.) | Mutation Type (prob.) | Elitisms |
|---|---|---|---|---|
| GA | 100 | Uniform (0.8) | Rand (0.08) | 0.03 |
| NSGA-II | 150 | Simulated Binary (0.9) | polynomial | 0.0 |

IV. RESULTS

The end result of this paper is a framework which is able to optimize humanoid's trajectory parameters for any given configuration. The framework consists of three nodes. The first one generates a CoM and Ankle Trajectory for any given input parameters and solves the inverse kinematics. The second node bridges between bullet physics engine and the optimization algorithm, which has been implemented in the last node. This node calculates cost functions for any solution requested by the optimization node. The connection between different nodes has been established using Robotic Operating System (ROS) communication protocols[1]. The schematic of the framework has been illustrated in Fig. 3. To evaluate the optimization solutions, we limited the search region for each of the optimization parameters. Ranges of the feasible region are given in Table II.

TABLE II.  SEARCH REGION OF OPTIMIZATION PARAMETERS

|  | $\alpha$ | $r_{DS}$ | $t_{step}(s)$ | $z_0(m)$ | $h_{ankle}(m)$ |
|---|---|---|---|---|---|
| min | 0.2 | 0.1 | 0.5 | 0.65 | 0.025 |
| max | 0.7 | 0.5 | 1.3 | 0.7 | 0.075 |

In the following, the obtained results will be reviewed.

*A. Single Objective Optimization*

First, we examined each of the four Objective Functions introduced in the previous section separately. The optimal parameters and the value of the cost function obtained for each of the cases can be found in Table III.

As mentioned before, there is a correlation between $J_E$, $J_{vel}$, and $J_{torque}$ Which could also be observed from optimal parameters resulted from GA. From Table III, it could be seen that $h_{Ankle}$ and $z_0$ for these three objective functions are similar and almost equal to the minimum and maximum values allowed for these parameters, respectively. It could be inferred from this outcome that the lower the height of the ankle and the higher the height of CoM get, the less energy is consumed, which is in accordance with our intuition. As seen and felt in our daily life, the more knee joint is bent, the more challenging walking becomes. Also, it could be seen that $t_{step}$ of ZMP objective function is quite smaller than that of other objective functions. Since walking speed is considered constant, this smaller value of $t_{step}$ results in a shorter step length and as we know the robot walks more stably when it takes shorter steps which approves our results.

---
[1] For more information about the implementation, you can refer to the following address:
https://github.com/CAST-Robotics/Trajectory-Optimization.

TABLE III. SINGLE OBJECTIVE OPTIMIZATION RESULTS

| $\alpha$ | $r_{DS}$ | $t_{step}$ | $z_0$ | $h_{ankle}$ | Objective Value |
|---|---|---|---|---|---|
| 0.42 | 0.34 | 1.09 | 0.696 | 0.026 | $J_E = 30.714$ KJ |
| 0.26 | 0.1 | 1.25 | 0.696 | 0.025 | $J_{torque} = 3348\ N.m$ |
| 0.48 | 0.33 | 1.19 | 0.699 | 0.025 | $J_{vel} = 155228\ rad/s$ |
| 0.38 | 0.1 | 0.6 | 0.658 | 0.036 | $J_{ZMP} = -132.54\ m$ |

However, optimal parameters' values associated with $J_{ZMP}$ rather differentiate from those of the other three objective functions, as shown in Table III. For the robot to walk most stably, $h_{Ankle}$ height has to be higher than when it walks in the most energy-efficient way, and its $\Delta z_{CoM}$ has to be lower. It is clear from these results that these two paradigms, walking with the most stability and walking most efficiently, contrast with each other. In other words, a humanoid cannot walk with the most stability margin while consuming the least energy possible. So a tradeoff must be made between these two demands, and that is where multi-objective optimization is helpful.

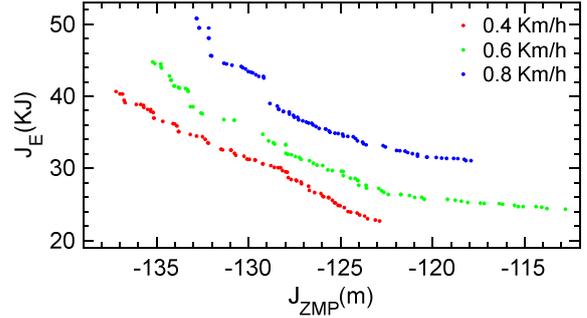

Fig. 4. Pareto set of the multi objective optimization with $J_E$ and $J_{ZMP}$ objective functions

### B. Multi-objective Optimization

We performed multi-objective optimization for energy, and ZMP cost functions at three different speeds: 0.4, 0.6, and 0.8 km / h. As the results in Fig. 4 show, both cost functions cannot be optimized simultaneously at all speeds, so the points in the middle of the graph are good choices for gait planning. In other words, there exists a solution that is closest to the knee point (also called utopia point) [24]. Also, Fig. 4 shows that energy consumption decreases as the robot slows down and the robot moves towards greater stability.

The optimal parameters for some of the points which have good results both in terms of energy and stability can be seen in Table IV. Also, with the help of the parameters at 0.6 Km/h, the trajectory of the Surena IV robot was designed, and the simulated robot followed it in the Choreonoid simulator. Fig. 5 shows a comparison between the designed and simulated trajectory. The designed trajectory is obtained from DCM algorithm, which is based on LIPM, while the simulated trajectory is the trajectory of the center of mass of robot's full dynamic model in simulation. To obtain the simulated trajectory, forward kinematics of the robot was solved in every iteration. Since we deployed the full dynamic model of the robot in simulation, we expected that these two trajectories would differ from each other, and this expectation is satisfied, as shown in Fig. 5.

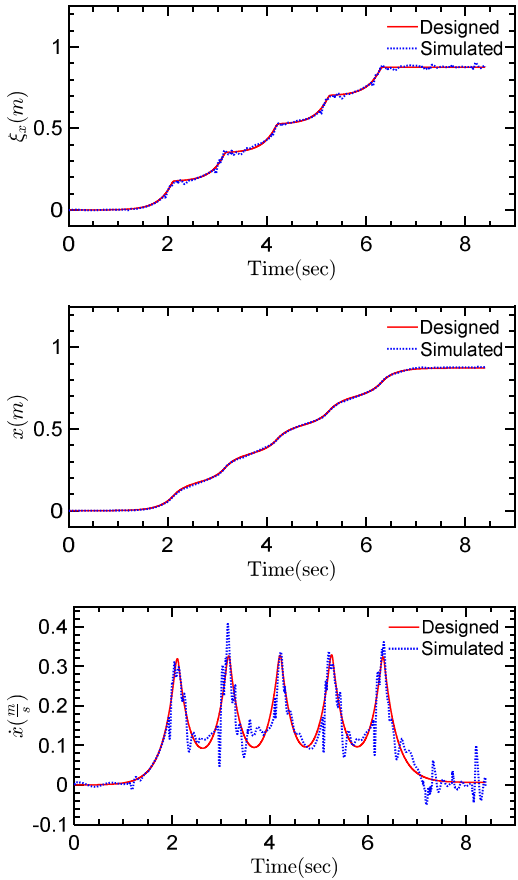
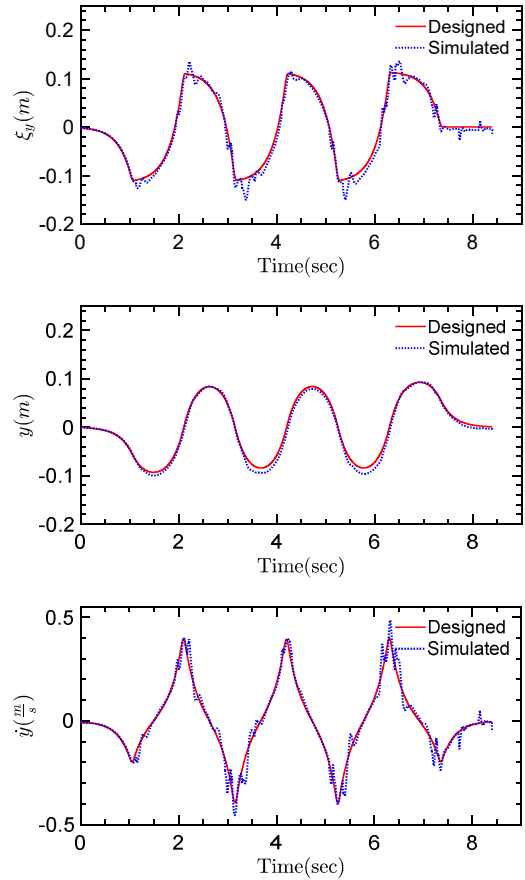

Fig. 5. Designed and simulated trajectories results for DCM, CoM, and CoM Velocity in both directions

TABLE IV. MULTI-OBJECTIVE OPTIMIZATION RESULTS

| speed ($Km/h$) | $\alpha$ | $r_{DS}$ | $t_{step}$ | $z_0$ | $h_{ankle}$ | Objective Values |
|---|---|---|---|---|---|---|
| 0.4 | 0.44 | 0.1 | 0.74 | 0.699 | 0.033 | $J_E = 31.175$ KJ $J_{ZMP} = -129.7\ m$ |
| 0.6 | 0.69 | 0.1 | 1.05 | 0.677 | 0.025 | $J_E = 32.079$ KJ $J_{ZMP} = -127.9\ m$ |
| 0.8 | 0.69 | 0.1 | 1.04 | 0.683 | 0.025 | $J_E = 36.278$ KJ $J_{ZMP} = -126.7\ m$ |

## V. CONCLUSION

In this paper, we developed a framework to obtain optimal parameters for trajectory planning based on DCM. The values of the cost functions in this optimization were obtained by considering the full dynamics model of the robot and the foot-ground contact model in the simulator. With the help of the obtained results, we can design the robot trajectory to move online with the most stability or the lowest energy consumption. It was also observed that it is not possible to achieve both of these goals simultaneously. Therefore, with the help of multi-objective optimization embedded in the framework, we obtained trajectory parameters that compromise these two objectives at three different speeds. We plan to design the optimal trajectory under different conditions, such as moving on uneven and slippery surfaces or soft surfaces for future works. As illustrated in Fig. 5, simulated trajectories during an offline walk has some fluctuations about the designed trajectory. With the help of online stabilizers, these fluctuations could be attenuated. Controlling robot's motion while rejecting output disturbances is another goal of the team for the future.